\documentclass{bmvc2k}

\usepackage{enumitem}
\usepackage{graphicx}
\usepackage{multirow}
\usepackage{booktabs}

\usepackage{tikz}
\usepackage{comment}
\usepackage{amsmath,amssymb} 
\usepackage{xcolor}
\usepackage{color,colortbl}
\input{insbox.tex}
\usepackage{wrapfig}
\usepackage{overpic}
\usepackage[normalem]{ulem}

\definecolor{ao}{rgb}{0.0, 0.0, 1.0}
\definecolor{airforceblue}{rgb}{0.36, 0.54, 0.66}
\definecolor{ceruleanblue}{rgb}{0.16, 0.32, 0.75}
\definecolor{cerulean}{rgb}{0.0, 0.48, 0.65}
\definecolor{celestialblue}{rgb}{0.29, 0.59, 0.82}
\definecolor{azure(colorwheel)}{rgb}{0.0, 0.5, 1.0}
\definecolor{babyblue}{rgb}{0.54, 0.81, 0.94}
\definecolor{babyblueeyes}{rgb}{0.63, 0.79, 0.95}
\definecolor{ballblue}{rgb}{0.13, 0.67, 0.8}

\definecolor{asparagus}{rgb}{0.53, 0.66, 0.42}
\definecolor{ao(english)}{rgb}{0.0, 0.5, 0.0}
\definecolor{applegreen}{rgb}{0.55, 0.71, 0.0}
\definecolor{armygreen}{rgb}{0.29, 0.33, 0.13}

\definecolor{amethyst}{rgb}{0.6, 0.4, 0.8}
\definecolor{antiquefuchsia}{rgb}{0.57, 0.36, 0.51}
\definecolor{blue-violet}{rgb}{0.54, 0.17, 0.89}
\definecolor{brightlavender}{rgb}{0.75, 0.58, 0.89}
\definecolor{brightube}{rgb}{0.82, 0.62, 0.91}
\definecolor{brilliantlavender}{rgb}{0.96, 0.73, 1.0}

\definecolor{amber}{rgb}{1.0, 0.75, 0.0}
\definecolor{amber(sae/ece)}{rgb}{1.0, 0.49, 0.0}
\definecolor{atomictangerine}{rgb}{1.0, 0.6, 0.4}
\definecolor{burntorange}{rgb}{0.8, 0.33, 0.0}
\definecolor{burntsienna}{rgb}{0.91, 0.45, 0.32}
\definecolor{cadmiumorange}{rgb}{0.93, 0.53, 0.18}
\definecolor{carrotorange}{rgb}{0.93, 0.57, 0.13}
\definecolor{chocolate(web)}{rgb}{0.82, 0.41, 0.12}
\definecolor{chromeyellow}{rgb}{1.0, 0.65, 0.0}
\definecolor{darkorange}{rgb}{1.0, 0.55, 0.0}
\definecolor{darktangerine}{rgb}{1.0, 0.66, 0.07}
\definecolor{deepcarrotorange}{rgb}{0.91, 0.41, 0.17}
\definecolor{deepsaffron}{rgb}{1.0, 0.6, 0.2}
\definecolor{fulvous}{rgb}{0.86, 0.52, 0.0}

\definecolor{tabred}{HTML}{d90429}
\definecolor{tabblue}{HTML}{fb8500}

\usepackage{tikz}
\usepackage{subcaption}
\usepackage{caption}
\usepackage{booktabs}
\newcommand{\modname}{\textsc{LPT}}
\graphicspath{{figures_lowres/}}

\usepackage[symbol]{footmisc}

\definecolor{C1}{HTML}{4363d8}
\definecolor{C2}{HTML}{f58231}
\definecolor{C3}{HTML}{3cb44b}
\definecolor{C4}{HTML}{e6194B}

\title{Learning to Construct 3D Building Wireframes from 3D Line Clouds}

\addauthor{Yicheng Luo$^{\text{\scriptsize 1,2}}$, Jing Ren$^{\text{\scriptsize 1,3}}$}{}{}
\addauthor{Xuefei Zhe$^{\text{\scriptsize 1}}$, Di Kang$^{\text{\scriptsize 1}}$}{}{}
\addauthor{Yajing Xu$^{\text{\scriptsize 2}}$, Peter Wonka$^{\text{\scriptsize 4}}$}{}{}
\addauthor{Linchao Bao}{}{1}

\addinstitution{
 Tencent AI Lab
}
\addinstitution{
 Beijing University of Posts and \\
 Telecommunications
}
\addinstitution{
 ETH Zurich
}
\addinstitution{
 KAUST
}
\runninghead{Luo}{Learning to Construct 3D Building Wireframes from 3D Line Clouds}

\begin{document}

\maketitle

\begin{abstract}
Line clouds, though under-investigated in the previous work, potentially encode  more compact structural information of buildings than point clouds extracted from multi-view images.
In this work, we propose the \emph{first} network to process line clouds for building wireframe abstraction.
The network takes a line cloud as input , i.e., a nonstructural and unordered set of 3D line segments extracted from multi-view images, and outputs a 3D wireframe of the underlying building, which consists of a sparse set of 3D junctions connected by line segments. 
We observe that a \emph{line patch}, i.e., a group of neighboring line segments,  encodes sufficient contour information to predict the existence and even the 3D position of a potential junction, as well as the likelihood of connectivity between two query junctions.
We therefore introduce a two-layer Line-Patch Transformer to extract junctions and connectivities from sampled line patches to form a 3D building wireframe model.
%
We also introduce a synthetic dataset of multi-view images with ground-truth 3D wireframe.
We extensively justify that our reconstructed 3D wireframe models significantly improve upon multiple baseline building reconstruction methods.The code and data can be found at \url{https://github.com/Luo1Cheng/LC2WF}.
\end{abstract}

\section{Introduction}
Recent advancement in photogrammetry makes it possible to obtain 3D data in city-scale from drone images.
Traditional point-based methods for 3D surface reconstruction from image such as multi-view stereo~\cite{campbell2008using,furukawa2009accurate,schonberger2016pixelwise} rely on accurate key point matching, which usually becomes challenging when facing texture-less surfaces (such as glass curtain) or large viewpoints changes.
%
%
To tackle this challenge, line segment-based methods have been proposed as a promising solution to camera pose estimation~\cite{miraldo2018minimal,salaun2016robustlinesfm} and surface reconstruction~\cite{surfaceLine,hofer2013incremental}.
It is shown to be easier and more robust to extract reliable line segments than points from multi-view images, especially in the case of lacking texture~\cite{Line3Dpp}.
%
%
%
Moreover, to alleviate computational costs of downstream geometry processing applications and to reduce storage cost of city-scale data, there is an increasing demand for urban reconstruction with \emph{lightweight} models such as 3D wireframe models or low-resolution polygonal meshes.
Besides, wireframe models are also widely-used in creating virtual cities or building information models.

\begin{figure}[!t]
    \centering
    \vspace{-2pt}
    \begin{overpic}[trim=0cm 0.6cm 0cm 0cm,clip,width=1\linewidth,grid=false]{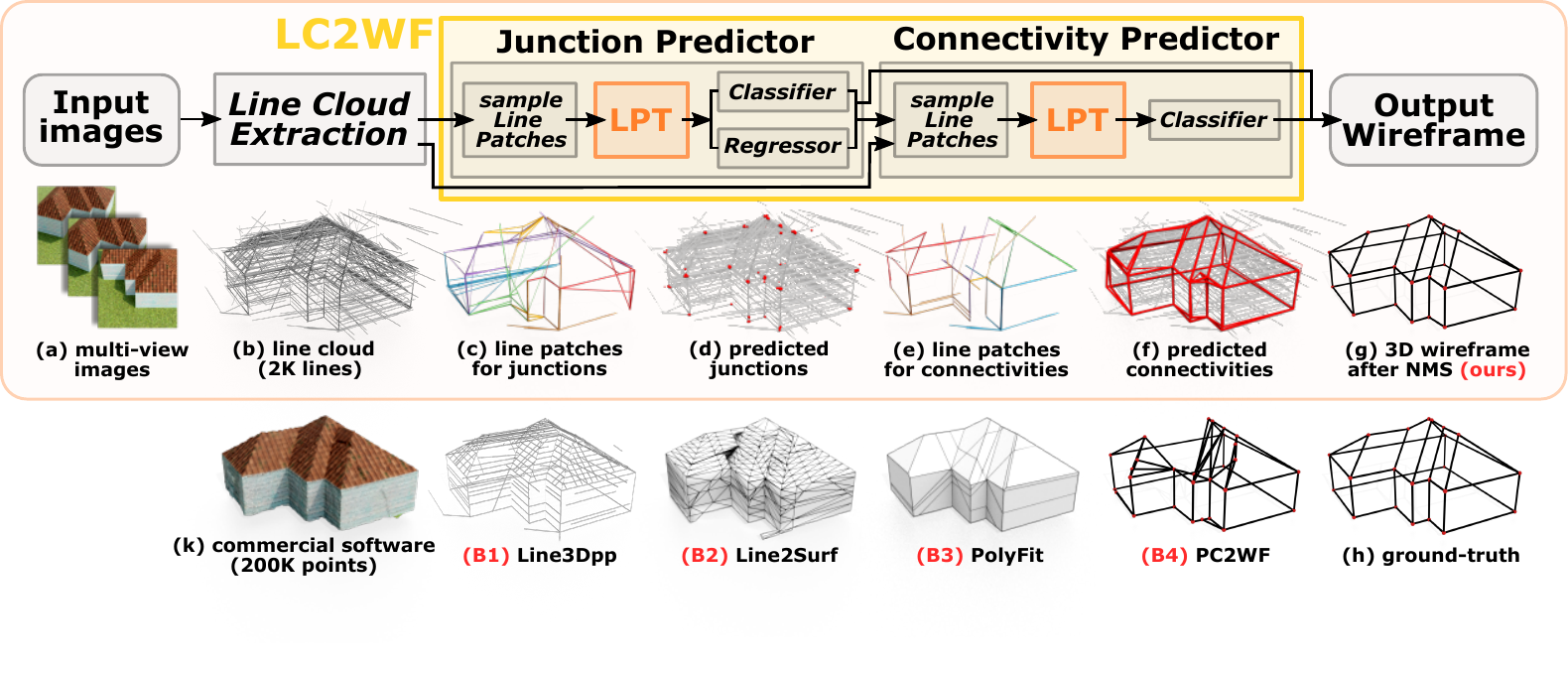}
    \end{overpic}\vspace{-2pt}
    \caption{\textbf{Method Overview.} \textbf{\emph{Top}}: our method takes multi-view images (a) as input and outputs a high-quality 3D wireframe (g). Specifically, we first extract a line cloud (b) from the images, from which we sample line patches (c) and (e) to predict wireframe junctions (d) and connectivities (f) respectively.
    \textbf{\emph{Bottom}}: we compare to four baselines: Line3Dpp~\cite{Line3Dpp} (B1) produces abstracted line clouds from the input noisy line clouds (b). Line2Surf~\cite{surfaceLine} (B2) takes (B1) as input and outputs a triangle mesh.
    From the point cloud (k), PolyFit~\cite{polyfit} (B3) produces a polygonal mesh, while PC2WF~\cite{PC2WF} (B4) outputs a 3D wireframe.}\label{fig:mtd_overview}\vspace{-18pt}
\end{figure}

To obtain lightweight building models, existing methods can be roughly categorized into two groups: (1) fit multiple simple primitives such as planes or boxes to the input point cloud to obtain a building abstraction as a polygonal mesh~\cite{polyfit,fangCVPR2020,lafarge2013surface,holzmann2017plane,holzmann2018semantically,vanegas2012automatic,li2016fitting,he2021manhattan}; (2) first construct a dense triangle mesh from the input point cloud using standard surface reconstruction techniques, e.g.~\cite{kazhdan2006poisson,labatut2009robustRecon,schonberger2016colmap}; then apply mesh simplification or decimation techniques to obtain an abstracted building model based on planar shape priors~\cite{chauve2010robustPlanar,salinas2015structure,bauchet2020kinetic,lafarge2012creating}.
However, both types of solutions rely on discrete operations (such as RANSAC-based fitting~\cite{polyfit} or region-growing for mesh decimation~\cite{salinas2015structure}), which makes it hard to adapt existing solutions for learning-based frameworks.
To close this gap, we present the \emph{first} learning-based solution for 3D building wireframe reconstruction.
We choose wireframe models as output since they are best suited for piece-wise planar objects such as urban buildings~\cite{PC2WF}. 
A wireframe is a graph representation of an object described by a set of junctions connected by line segments. 
Wireframe models have become popular for characterizing the contours of objects in both 2D~\cite{LCNN,HAWP,PPGNET,Kong2021hole} and 3D~\cite{PC2WF}. 
However, learning a 3D building wireframe from a point or line cloud is a challenging and under-explored task, which still remains an open problem.

In this work, we propose a solution to extract the 3D building wireframe from a line cloud. As observed in~\cite{holzmann2018semantically,he2021manhattan}, line clouds potentially provide more structural information such as corner points and boundary edges of buildings, which are much harder to extract from point clouds. 
Moreover, a line cloud is more compact to characterize a building than a point cloud. 
For example, our method can output a reasonable building wireframe from a line cloud containing around 1K line segments. To achieve comparable result, a dense point cloud containing 50K-100K points is required for baseline methods such as PolyFit~\cite{polyfit}.

To summarize, our main \emph{contributions} are: 
\textbf{(1)} a novel \emph{learning-based} solution to reconstruct 3D building wireframe from multi-view images; 
\textbf{(2)} LC2FW: a transformer-based and the \emph{first} network to process line clouds based on line patches;
\textbf{(3)} an adapted synthetic dataset with annotated multi-view images and ground-truth 3D wireframe models. 

\vspace{-15pt}
\section{Related Work}
There is relatively limited work that focuses on building wireframe reconstruction from either multi-view images or point clouds. We mainly review related work of building reconstruction, wireframe reconstruction, as well as existing datasets for building reconstruction.

\noindent{\textbf{3D Point/Line Reconstruction}}
Structure-from-Motion~\cite{snavely2006photo,schonberger2016colmap,agarwal2011buildingRome,crandall2011discrete,snavely2008skeletal,sweeney2015optimizing} is an effective method to acquire 3D point clouds or line clouds for surface reconstruction from multi-view images. 
Corresponding feature points extracted from multi-view images are used to estimate camera parameters and generate 3D point clouds. 
Similarly, 3D line clouds can be generated from corresponding 2D line segments detected from multi-view images~\cite{Line3Dpp}. 
In our work, we focus on line clouds since the building shapes can be easily characterized by line structures.

\vspace{-15pt}
\paragraph{\textbf{Building Reconstruction}}
Multiple optimization-based algorithms have been proposed for building reconstruction from point clouds.
Some works~\cite{he2021manhattan,manhattan2016,li2016fitting,3Dwireframe} use the Manhattan-world assumption to further regularize the building reconstruction.
%
%
The reconstructed building meshes are usually dense and noisy, and thus different methods have been proposed for simplification or abstraction~\cite{verdie2015lod,BigSur,li2021feature}.
%
%
Primitive-based building reconstruction is another popular direction to get abstracted polygonal mesh by exploiting high-level primitives such as cubes~\cite{vanegas2012automatic,li2016fitting,he2021manhattan}, planes~\cite{polyfit,fangCVPR2020,lafarge2013surface,holzmann2017plane,holzmann2018semantically}, or general 3D templates~\cite{nan2015template,lin2013semantic} to fit input point clouds of buildings.
%
%
However, building reconstruction from a 3D \emph{line cloud} has been rarely investigated.
Existing works~\cite{sugiura20153d,holzmann2018semantically,surfaceLine,he2021manhattan} take 3D lines into consideration to fit planes first, instead of directly extracting the building structure from the lines.
Sugiura et al.~\cite{sugiura20153d} extend the tetrahedra-carving method to the 3D point-and-line cloud setting, while Holzmann et al.~\cite{holzmann2018semantically} use additional semantic labels from image segmentation to cluster lines for plane fitting. 
Langlois et al.~\cite{surfaceLine} propose a RANSAC-based method to extract planes from the input line cloud, which are fused to form a watertight mesh.
He et al.~\cite{he2021manhattan} estimate planes and corners from a line cloud for box fitting.
Some other works~\cite{hofer2015line3d,Line3Dpp} provide heuristics for line cloud abstraction.
In our work, we propose the first \emph{learning-based} solution to process line clouds for 3D building wireframe reconstruction.

\noindent \textbf{Wireframe Extraction}
As a special case of 2D edge detection~\cite{AFM,semanticLine,HT-HAWP,LETR,TP-LSD,ELSD,SOLD2,LSD,gu2021line},
2D wireframe detection from a single image~\cite{LCNN,HAWP,PPGNET,Kong2021hole} is much more explored compared to the 3D wireframe reconstruction setting.
%
%
A recent work PC2WF~\cite{PC2WF} proposes a CNN-based method to extract 3D wireframe models from point clouds.
Zhou et al.~\cite{3Dwireframe} provide a method to reconstruct \emph{partial} 3D wireframe models from a single image, from which depth maps, junction heatmaps, edge maps, and vanishing points are estimated independently for wireframe prediction.
%
%
In our work, a \emph{complete} 3D wireframe model is reconstructed from a \emph{noisy} line cloud extracted from multi-view images.

\vspace{-15pt}
\paragraph{\textbf{Dataset}}
There are multiple datasets that contain ground-truth 2D lines in images with semantically meaningful annotations~\cite{semanticLine,Lee2017SLNet}.
Here we mainly review datasets that can be potentially used for either wireframe or building reconstruction.
\cite{wireframeDataset} and~\cite{YorkUrbanDataset} provide ground-truth 2D wireframe annotations for single images of indoor or outdoor scenes.
\cite{3Dwireframe} proposes a synthetic city dataset that contains 2D synthetic images with ground-truth depth and partial 3D wireframe annotations that are visible from a single view.
There are also some datasets consisting of CAD models~\cite{abcdataset} or polygonal meshes~\cite{ren2021intuitive} that can be potentially adapted to wireframes.
The ABC dataset~\cite{abcdataset} is a recent dataset consisting of one million CAD models, most of which are mechanical parts.
In this work, we build on~\cite{ren2021intuitive} to create a synthetic dataset with \emph{complete} ground-truth 3D wireframe models paired with multi-view images.

\vspace{-10pt}
\section{Background \& Training Dataset}

\noindent \textbf{Notation} Our method takes a set of multi-view images $\mathcal{I} = \{I_i\}_{i=1}^{m}$ as input, from which we extract a \emph{line cloud}~\cite{Line3Dpp,he2021manhattan} that consists of a group of \emph{line segments} $\mathcal{L} = \{l_i\}_{i=1}^{N}$, where each line segment $l_i$ is denoted by its two 3D endpoints, i.e., $l_i = (p_i, q_i), p_i, q_i \in\mathbb{R}^3$.
We denote $\mathcal{G}$ as a group of line segments belonging to $\mathcal{L}$, i.e., $\mathcal{G}\subset \mathcal{L}$. 
The underlying 3D \emph{wireframe} model of the line cloud $\mathcal{L}$ is denoted as $\mathcal{W} = \big( \mathcal{V}, \mathcal{E} \big)$, which is defined by a set of 3D vertices (junctions) $\mathcal{V}$ and a set of edges (connectivities) $\mathcal{E}$ that connect those vertices. Specifically, we have $\mathcal{V} = \{v_i\}_{i=1}^{n_v}, v_i\in\mathbb{R}^3$, and $\mathcal{E} \subset \mathcal{V}\times\mathcal{V}.$

\noindent \textbf{Overall Pipeline}
%
The goal of our method is to output an accurate and clean wireframe model $\mathcal{W}$ from a set of input images $\mathcal{I}$ of a building. 
Our method contains the following major building blocks (see Fig.~\ref{fig:mtd_overview}): (1) a \emph{line cloud extraction} step where a dense line cloud is extracted from the input images (Sec.~\ref{sec:mtd:line_cloud}); 
(2) a \emph{junction predictor} which classifies if there exists a junction in a group of lines and regresses the junction position accordingly (Sec.~\ref{sec:mtd:junc_pred}); 
(3) a \emph{connectivity predictor} that instantiates edges between the predicted junctions (Sec.~\ref{sec:mtd:edge_pred}).


\vspace{-10pt}
\subsection{Line Cloud Extraction}\label{sec:mtd:line_cloud}
\setlength{\columnsep}{2pt}%
\setlength{\intextsep}{0pt}%
\begin{wrapfigure}{r}{0.25\linewidth}
\centering
\begin{overpic}[trim=0.2cm 0cm 0.6cm 0.6cm,clip,width=1\linewidth,grid=false]{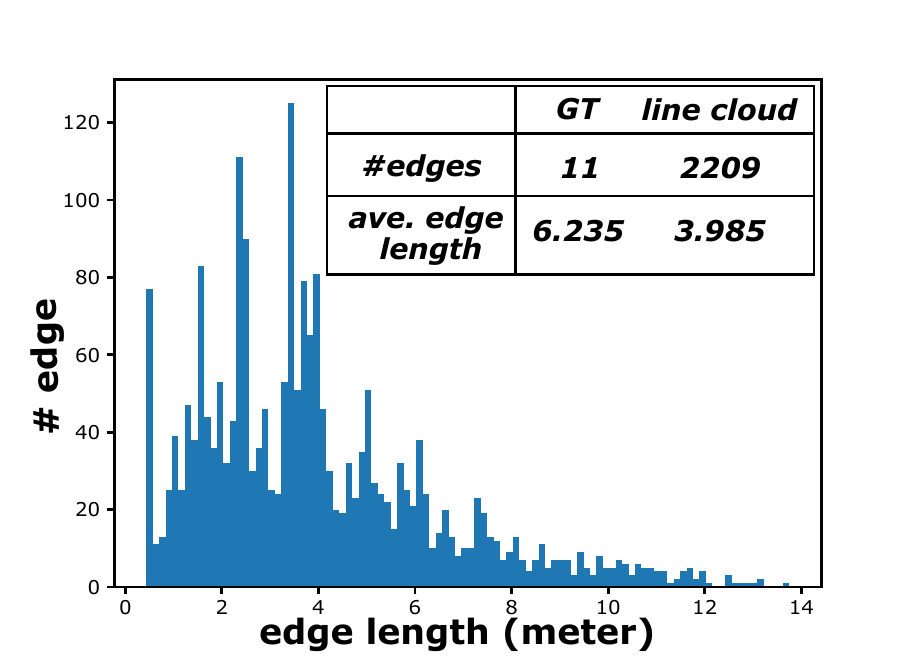}
\end{overpic}
\end{wrapfigure}

There are roughly two groups of methods to extract a line cloud from multi-view images: 
(1) reconstruct 3D lines and estimate camera parameters simultaneously~\cite{PLVIO,lineBasedSLAM,salaun2017line}; 
(2) reconstruct 3D lines with fixed camera parameters estimated from standard structure-from-motion (SfM) methods~\cite{Line3Dpp,jain2010exploiting}.
In the work, we follow the latter one to reconstruct a line cloud, which is also adopted in Line3Dpp~\cite{Line3Dpp}, the current state-of-the-art line cloud abstraction method.
Specifically, the camera parameters are estimated from the multi-view images using SfM.
Correspondences between the 2D line segments detected from each image (using any existing line detector) are established based on epipolar constraints, which are then used to solve 3D line segments based on the camera parameters.
We use the line cloud extractor as provided in~\cite{Line3Dpp}.
Note that, the extracted line cloud is potentially dense, noisy, and incomplete. 
The inset figure shows the histogram of the length of the line segments in the line cloud shown in Fig.~\ref{fig:mtd_overview} (b). Around 85\% of the extracted line segments has a shorter length than the average edge length of the underlying building (Fig.~\ref{fig:mtd_overview} (h)). This suggests that the extracted line clouds contain large portion of short (and potentially noisy in orientations) line segments, which makes it challenging to extract a clean wireframe.
\vspace{-10pt}
\subsection{BuildingWF Dataset: Training Dataset}\label{sec:mtd:dataset}

\noindent \textbf{Challenges}
To design a data-driven solution for building wireframe reconstruction, we need large-scale datasets with ground-truth 3D wireframe annotations paired with either multi-view images or point clouds. However, it is quite challenging to obtain such datasets.
Existing building datasets can be roughly categorized as follows: (1) single image with ground-truth 2D line segments~\cite{semanticLine}; (2) single image with ground-truth 2D wireframe~\cite{wireframeDataset,YorkUrbanDataset}; (3) single depth image with ground-truth \emph{partial} 3D wireframe that is visible in the image~\cite{3Dwireframe}.
On the other hand, the dataset used in PC2WF~\cite{PC2WF} are indeed in large-scale but only contain ground-truth 3D wireframe for \emph{man-made objects} such as mechanical objects~\cite{abcdataset} and furniture.


\noindent \textbf{BuildingWF Dataset} 
In this work, we introduce a synthetic dataset with ground-truth 3D building wireframe models based on the 
Roof-Image dataset proposed in~\cite{ren2021intuitive}, which contains around 3.6K polygon meshes of residential buildings (denoted as $\mathcal{M}_i$).  See supplementary materials for some examples.
We first extract the ground-truth wireframe $\mathcal{W}^{\text{gt}} = (\mathcal{V}^{\text{gt}}, \mathcal{E}^{\text{gt}})$ from the provided building mesh $\mathcal{M}$.
We then synthesize multi-view images $\mathcal{I}$ of the building $\mathcal{M}$ in Blender with synthetic textures based on the provided face labels.
A 3D line cloud $\mathcal{L}$ is extracted from synthetic images $\mathcal{I}$ as mentioned in Sec.~\ref{sec:mtd:line_cloud}.
We then use the ground-truth wireframe $\mathcal{W}^{\text{gt}}$ and camera parameters to label each 3D line segment in the line cloud $\mathcal{L}$. 

Specifically, we first project the 3D ground-truth wireframe $\mathcal{W}^{\text{gt}}$ to image planes using the corresponding camera parameters to get the ground-truth 2D wireframe for \emph{each} image $I_i\in\mathcal{I}$, which allows us to check if a 3D line $l_i\in\mathcal{L}$ is part of the wireframe $\mathcal{W}^{\text{gt}}$ or not.
If the 2D line segments, that are used to reconstruct the 3D line $l_i$, are close enough to the ground-truth 2D wireframes, $l_i$ will be classified as part of $\mathcal{W}^{\text{gt}}$ and be labeled as 1.
For a line $l_i$ with label 1, we further associate it with two ground-truth junction vertices that are the endpoints of the corresponding edge in $\mathcal{W}^{\text{gt}}$ that $l_i$ belongs to.
In summary, each line $l_i$ has a 5-dimensional label: $(f, i_1, d_1, i_2, d_2)$ where $f$ is binary indicating if this line is part of the wireframe, $i_1, i_2$ are the junction index in $\mathcal{V}^{\text{gt}}$ and $d_1, d_2$ are the distances from $l_i$ to the two ground-truth junctions respectively if $f = 1$.
Note that the label $f$ is used to supervise our junction classifier, and the remaining labels are used to supervise our junction regressor.


\vspace{-12pt}
\section{LC2WF: Line Cloud to Wireframe}\label{sec:lc2wf}
In this section, we present the key component of our framework, LC2WF network that reconstructs a 3D wireframe from a line cloud.
Before we dive into the architecture details, we would like to first motivate our design choices.
The core problem is how to correctly predict the junction positions and the connectivities between junctions from a line cloud.
Similar to a point cloud, a line cloud is \emph{nonstructural}, \emph{dense}, \emph{noisy}, and potentially \emph{incomplete}. 
However, on the other hand, the \emph{orientation} and \emph{length} is properly defined for line segments, which makes the neighborhood in a line cloud potentially more informative than the neighborhood in a point cloud, where only the distance between points is defined.

In the following, we first introduce \emph{line patches} to define the neighborhood in a line cloud. We then propose our line-patch transformer\cite{transformer}, LPT, that is designed to process line patches to effectively extract information for junction and connectivity prediction, which are combined to produce the final 3D wireframe.

\begin{figure}[!t]
    \centering\vspace{-3pt}
        \begin{overpic}[trim=0.8cm 1.2cm 0.2cm 0.5cm,clip,width=1\linewidth,grid=false]{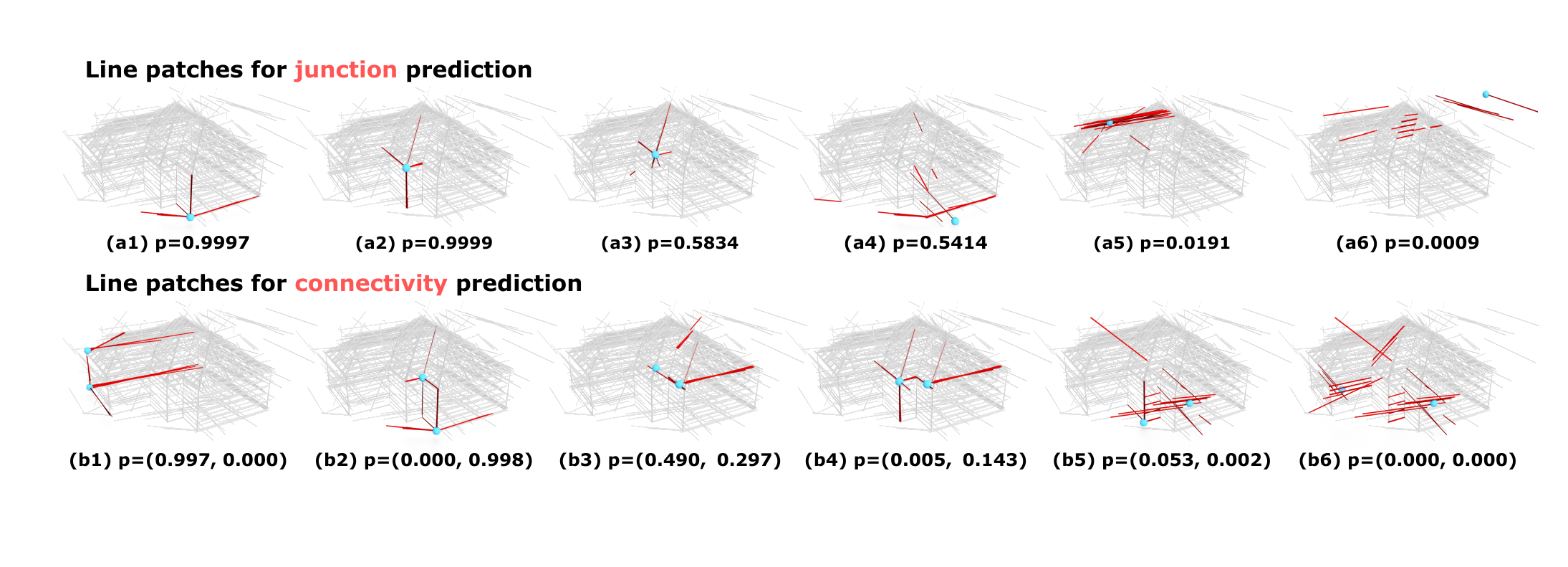}
    \end{overpic}\vspace{5pt}
    \caption{Example line patches (red lines) w.r.t. the sampling points (blue). \emph{\textbf{Top}}: we report the probability for each line patch to have a junction. \emph{\textbf{Bottom}}: we report two probabilities for a pair of sampling points, i.e., (1) two points are connected, and (2) two points that potentially have graph distance of 2. Note all the line patches have the same number of lines.}\label{fig:mtd:eg_line_patches}\vspace{-9pt}
\end{figure}
\vspace{-10pt}
\subsection{Line Patches}\label{sec:mtd:line_patch}
In our setting, a line patch is defined as a group of 3D line segments collected w.r.t. sampling points. Specifically, given an arbitrary point $x\in\mathbb{R}^{3}$, the corresponding line patch $\mathcal{G}(x)$ is defined as:
$\mathcal{G}(x) = \big\{ l\in \mathcal{L} \,\vert\, \text{dist}(x, l) \le \epsilon \big\}$, 
where $\text{dist}(x, l)$ measures the \emph{point-to-line} distance between a point $x$ and the 3D line where the line segment $l$ lies. We can similarly define the line patch w.r.t. a pair of sampling points as $\mathcal{G}(x,y) = \mathcal{G}(x) \cup \mathcal{G}(y)$. 
We observe that the line patches encode sufficient information to predict junction positions and connectivity between junctions. Specifically, the line patch $\mathcal{G}(x)$ of point $x$ can be used to estimate the probability of having a junction located around point $x$, while the line patch $\mathcal{G}(x, y)$ can be used to estimate the probability of having an edge connecting the point $x$ and point $y$.

See Fig.~\ref{fig:mtd:eg_line_patches} for an illustration: 
In example (a1) and (a2), the lines in the red patch have multiple dominant orientations, suggesting that the blue sampling point is indeed close to a wireframe junction. This aligns with the fact that a 3D junction is formed by the intersection of at least three planes, and the corresponding intersecting lines shape the contour of the underlying building, which would lead to dominant line clusters in images.
The blue sampling point in example (a5) is located on a roof plane, where the roof texture (see Fig.~\ref{fig:mtd_overview}) contains structural lines. In this case there exists only one dominant direction, which is not enough to support a junction.
Example (a6) shows the line patch of an outlier point, where the lines in the patch are extremely unstructured. 
Similar for the examples in (b1-b6), we can see that if two sampling points are likely to be connected to each other, the corresponding line patch will reveal strong pattern (e.g., having duplicated lines) to support it.
All these observations of line patches perfectly align with the properties of the wireframe models of planar objects: the junctions are formed by the intersection of planes with at least three dominant directions determined by the intersecting lines. As a comparison, other points such as corners in textures or noisy points do not have comparably strong signals.

\vspace{-10pt}
\subsection{Line-Patch Transformer (LPT)}\label{sec:mtd:lpt}
Given a line patch $\mathcal{G}(x)$ or $\mathcal{G}(x,y)$, how can we tell if there exists a junction or an edge?
We propose a line-patch transformer, {\modname}, to extract features from line patches, which can then be used to predict the junctions/edges.
Specifically, a line patch $\mathcal{G}(x)$ can be represented as a 2D tensor $(N, F^{\text{in}})$, that stores $N$ neighboring lines in $\mathcal{G}(x)$, and each line has $F^{\text{in}}$ features including the coordinates of the two endpoints and the distance %
between the line and the
\setlength{\columnsep}{1pt}%
\setlength{\intextsep}{1pt}%
\begin{wrapfigure}{r}{0.15\linewidth}
\centering
\begin{overpic}[trim=-0.2cm 0cm 0cm 0cm,clip,width=1\linewidth,grid=false]{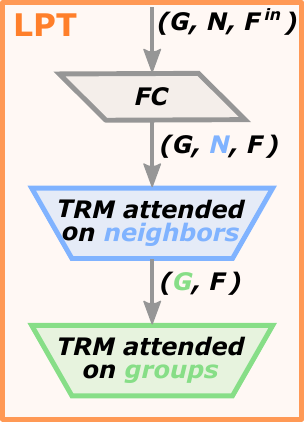}
\end{overpic}\vspace{-8pt}
\end{wrapfigure}
sampling point $x$.
We then collect $G$ groups of line patches in a 3D tensor $(G, N, F^{\text{in}})$. 
{\modname} contains two transformers (see inset figure): (1) the first transformer attends to the $N$ neighbors for each line patch, to potentially find the most prominent lines for junction predictions; (2) the second transformer attends on the $G$ groups, to potentially attend to the junctions that are co-planar. 
%
We believe the attention mechanism inside the transformers is more effective in capturing \emph{local} geometry information in line patches as well as the \emph{global} geometry in the line cloud.
We similarly use {\modname} to process line patches $\mathcal{G}(x,y)$ for connectivity prediction, where the initial features can be obtained by concatenating the features of $\mathcal{G}(x)$ and $\mathcal{G}(y)$.

\vspace{-10pt}
\subsection{Junction Predictor}\label{sec:mtd:junc_pred}
We sample $G$ points $\{x_k\}_{k=1}^{G}$ from all the line endpoints of the line cloud $\mathcal{L}$ to construct line patches for junction prediction. 
Specifically, we first sample a smaller set of points (around 25\%) according to the endpoint density and then sample the remaining points via Farthest Point Sampling (FPS)~\cite{fps}. 
We then obtain the corresponding line patch $\mathcal{G}(x_k)$ for each sample $x_k$ as discussed in Sec.~\ref{sec:mtd:line_patch}.
The line patches $\{ \mathcal{G}(x_k) \}$ are fed into LPT to extract patch features, which are used to \emph{classify} if there exist a junction close to $x_k$, and \emph{regress} the potential junction position $p_k$. We then collect the predicted junctions $p_k$ in $\mathcal{V}^{\text{pred}}$.
The classifier can help to filter out junctions with a low confidence. 
During the training phase, we first draw samples that are close to the ground-truth junctions, then sample via density and FPS to get $G$ sampling points for constructing the line patches. 
This guarantees that we draw both positive line patches (containing a junction) and negative line patches.
Specifically, a line patch is considered as a negative sample if there are more than half of the line segments in the patch are labeled as noise (introduced in Sec.~\ref{sec:mtd:dataset}).
The loss function for the classifier is a binary cross-entropy $E_{\text{v-clf}}$.
The loss function for the regression is L$_2$ distance between the predicted position and the ground-truth position.

\vspace{-10pt}
\subsection{Connectivity Predictor}\label{sec:mtd:edge_pred}
We first sample $G$ pairs of predicted junctions $(p_k, q_k)\in \mathcal{V}^{\text{pred}}\times \mathcal{V}^{\text{pred}}$ w.r.t. the predicted probability.
We can then construct the line patches $\{\mathcal{G}(p_k, q_k)\}$ and feed them into LPT to extract patch features, which is used to classify the junction pair $(p_k, q_k)$ into five groups: (A) labeled as -1 if at least one of $(p_k, q_k)$ is a false positive junction (i.e., does not belong to the underlying wireframe); (B) two vertices are true positive junctions and the pair is labeled by the \emph{graph distance} in the underlying wireframe, i.e., (B.0) with graph distance 0 ($p_k$ is identical to $q_k$), (B.1) with graph distance 1 ($p_k$ is connected to $q_k$), (B.2) with graph distance 2 ($p_k, q_k$ are adjacent to the same vertex), or (B.3) having graph distance larger than 2.
The proposed fine-grained classification with 5 categories can provide more \emph{informative} labels and more \emph{balanced} distribution (which leads to better results as justified in Tab.6 in the supplementary) than a naive binary classification, which implicitly assumes all predicted junctions are true positive and are not redundant.
Note that the edge labels can help to further prune the false positive junctions besides the probability produced by the junction classifier.
During training, we sample from $\mathcal{V}^{\text{gt}}\times\mathcal{V}^{\text{gt}}$ and $\mathcal{V}^{\text{pred}}\times\mathcal{V}^{\text{pred}}$ to learn junction connectivity.
A vertex from $\mathcal{V}^{\text{pred}}$ is regarded as a false positive junction if its distance to the nearest ground-truth junction is larger than a threshold $\epsilon$. Any vertex pair that contains a false positive junction is labeled as -1. The rest vertex pairs is labeled according to the graph distance in the ground-truth wireframe, where for a pair of predicted junctions, we use the graph distance between their nearest ground-truth junctions. The loss function for the classifier is standard cross-entropy $E_{e\text{-clf}}$. 

\vspace{-10pt}
\subsection{Implementation Details}
\noindent \textbf{Network Details}
Our \emph{LPT architecture} includes fully-connected (FC) layers (with batch normalization and ReLU activation) and two transformer encoder layers (with layer normalization, ReLU activation, and pre-normalization). The output sizes of the FC layers are set to 64/64/128/128/256 resp. In the transformer encoder layer, the input size and the latent layer size are set to 256.
The \emph{classifier/regressor} for junction prediction, and the \emph{classifier} for connectivity prediction similarly include FC layers, ReLU activation, and batch normalization. The output size of the FC layers are 256, 128, 64, 32, and 2/3/5 (for junction-classifier/junction-regressor/connectivity-classifier resp.). See the supplementary for more details.

\noindent\textbf{Training Loss}
The total training loss for our complete networks is: $E_{\text{total}} = E_{v\text{-clf}} + \lambda_v E_{v\text{-reg}} + \lambda_e E_{e\text{-clf}}$, where $\lambda_v, \lambda_e$ are balancing weights.

\noindent \textbf{Post-processing}
For post-processing, we first use non-maximum suppression (NMS) to remove duplicated vertices and redundant edges that are close to each other. We then use the connectivity predictor to further prune the predicted junctions that tend to be false positives. Specifically, if a vertex pair is categorized to be identical (i.e., with label 0), then the junction with a lower confidence will be removed. For two vertices with similar Hamming distance in adjacency and small Euclidean distance, the vertex with a lower confidence will be removed. We also remove the isolated edges from the final wireframe.

\vspace{-20pt}
\section{Experiments}
We compare different methods for building mesh/wireframe reconstruction on our BuildingWF dataset with ground-truth annotations (introduced in Sec.~\ref{sec:mtd:dataset}). We briefly introduce the baselines  and the metrics for evaluation in Sec.~\ref{sec:res:baselines}.
In Sec.~\ref{sec:res:comparison} we show quantitative and qualitative results on building wireframe reconstruction.
See supplementary materials for ablation study, more results and discussions. Code and data will be released.
\vspace{-10pt}
\subsection{Baselines \& Evaluation Metrics}\label{sec:res:baselines}
\setlength{\columnsep}{1pt}%
\setlength{\intextsep}{1pt}%
\begin{wraptable}{r}{0.3\linewidth}
    \caption{Baselines}\label{tb:res:baseline}\vspace{3pt}
    \centering
\tiny
{\def\arraystretch{1}\tabcolsep=0.5em 
\begin{tabular}{lcccc}
\toprule[0.8pt]
\multicolumn{1}{c}{Method} & \begin{tabular}[c]{@{}c@{}}Input\\ type\end{tabular} & \begin{tabular}[c]{@{}c@{}}Input\\ size\end{tabular} & \begin{tabular}[c]{@{}c@{}}Output\\ type\end{tabular} & \begin{tabular}[c]{@{}c@{}}Runtime\\ (sec)\end{tabular} \\ \midrule[0.8pt]
line3Dpp & \cellcolor{babyblue!15}lines & 1,388 & \cellcolor{babyblue!15}lines & 33.1 \\
line2Surf. & \cellcolor{babyblue!15}lines & 120 & \cellcolor{babyblue!15}mesh & 220.8 \\
PolyFit & \cellcolor{babyblue!15}points & 86,396 & \cellcolor{babyblue!15}mesh & 45.6 \\  \midrule[0.5pt]
PC2WF & \cellcolor{babyblue!15}points & 86,396 & \cellcolor{babyblue!15}wireframe & 31.7 \\ 
Ours & \cellcolor{babyblue!15}lines & 1,388 & \cellcolor{babyblue!15}wireframe & 0.9 \\\bottomrule[0.8pt]
\end{tabular}
}


\end{wraptable}
To the best of our knowledge, there is no existing baseline for reconstructing building wireframes from multi-view images directly. We therefore mainly compare to the state-of-the-art 3D line cloud abstraction method, line3Dpp~\cite{Line3Dpp}, building reconstruction methods, Line2Surface~\cite{surfaceLine} and PolyFit~\cite{polyfit}, and 3D wireframe reconstruction method PC2WF~\cite{PC2WF}.
Specifically, line3Dpp~\cite{Line3Dpp} outputs an abstracted line cloud from a dense line cloud based on heuristics for line clustering.
Line2Surface~\cite{surfaceLine} is an optimization-based method that extracts planes from a line cloud via RANSAC to form a building mesh. 
PolyFit~\cite{polyfit} is the state-of-the-art optimization-based method for building mesh reconstruction from a potentially noisy point cloud. 
PC2WF~\cite{PC2WF} is a novel learning-based method to reconstruct a 3D wireframe from a point cloud, which achieves plausible results on man-made objects such as mechanical objects and furniture.
For evaluation, we follow PC2WF\cite{PC2WF} to measure the precision and recall on both predicted junctions and wireframes, and the Wireframe Edit Distance(WED): 
(1) $v\text{AP}_{\eta}$ and $v\text{Recall}_{\eta}$ show the precision/recall on the predicted \emph{junctions}.
(2) $s\text{AP}_{\eta}$ and $s\text{Recall}_{\eta}$ report the \emph{structural} quality of the predicted \emph{wireframes}. Specifically, it checks if a predicted edge is a true positive or if a ground-truth edge is retrieved according to the distances between the edge endpoints.
(3) WED reports the number of operations and the editing distances of adding/removing predicted junctions/edges that are needed to transform the graph structure of the predicted wireframe into the ground-truth wireframe.

\vspace{-10pt}
\subsection{Results and Comparisons}\label{sec:res:comparison}
\begin{figure}[!t]
    \centering
    \begin{overpic}[trim=0.5cm 0.8cm 0.15cm 0.4cm,clip,width=1\linewidth,grid=false]{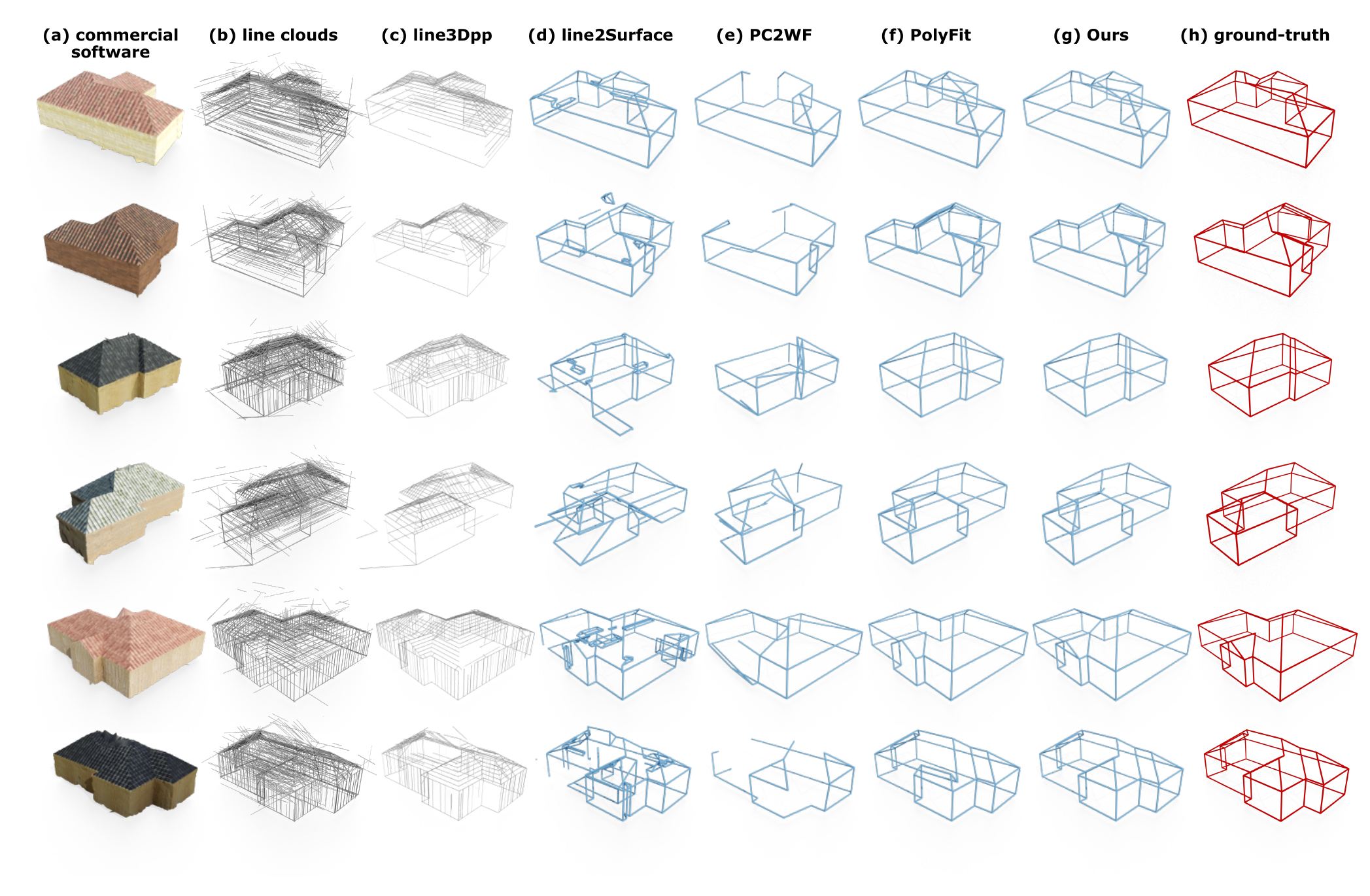}
    \end{overpic}\vspace{3pt}
    \caption{Comparison to baselines on building wireframe reconstruction.}
    \label{fig:res:baselines}\vspace{-15pt}
\end{figure}

We compare to baseline methods on 757 test buildings.
The line clouds (for line3Dpp, line2Surf, and our method) and the point clouds (for PolyFit and PC2WF) are extracted using the \emph{same} camera parameters. Note that we use a \emph{commercial software} to extract high-quality point clouds (See Fig.~\ref{fig:mtd_overview} (k) and Fig.~\ref{fig:res:baselines} (a) for some examples). 
Moreover, to make a fair comparison to line2Surface~\cite{surfaceLine} and PolyFit~\cite{polyfit}, we post-process the output meshes into wireframes by merging co-planar faces and parallel adjacent edges, removing interior edges and isolated vertices, etc. For PC2WF we use the provided NMS for post-processing. %
We report the evaluations on the results after post-processing (Tab.~\ref{tab:res:PRandWED}).


\begin{table}[!t]
    \begin{subtable}[t]{0.53\textwidth}
    \caption{Precision/Recall of the predicted \textbf{junctions} ($v\text{AP}$/$v\text{Recall}$) and the predicted \textbf{wireframe} models ($s\text{AP}$/$s\text{Recall}$) on results \textbf{after post-processing}. We highlight the \textcolor{tabred}{\textbf{best}} and the \textcolor{tabblue}{\textbf{second best}} results.}\label{tab:res:postprocess}
    \centering
\tiny
{\def\arraystretch{1.05}\tabcolsep=0.3em
\begin{tabular}{lcccccccc}
\toprule[0.8pt]
& \multicolumn{4}{c}{$v\text{AP}_{\eta}$/$v\text{Recall}_{\eta}$  (\%)} & \multicolumn{4}{c}{$s\text{AP}_{\eta}$/$s\text{Recall}_{\eta}$ (\%)} \\ \cmidrule(lr){2-5}\cmidrule(lr){6-9} 
\multicolumn{1}{c}{\multirow{-2}{*}{Method}} & $\eta=0.15$ & $\eta=0.25$ & $\eta=0.35$ & avg. & $\eta=0.25$ & $\eta=0.35$ & $\eta=0.50$ & avg. \\ \midrule[0.8pt]
line2Surf.  & 26.7/\textcolor{tabblue}{83.9} & \cellcolor{babyblue!15}27.4/\textcolor{tabblue}{85.8} & 27.6/86.6 & \cellcolor{babyblue!15}27.2/\textcolor{tabblue}{85.4} & 24.2/\textcolor{tabblue}{58.8} & \cellcolor{babyblue!15}25.1/61.0 & 25.8/62.6 & \cellcolor{babyblue!15}25.0/60.8 \\
PolyFit  & \textcolor{tabblue}{52.1}/70.8 & \cellcolor{babyblue!15}\textcolor{tabblue}{62.0}/84.3 & \textcolor{tabblue}{64.3}/\textcolor{tabblue}{87.4} & \cellcolor{babyblue!15}\textcolor{tabblue}{59.5}/80.8 & \textcolor{tabblue}{45.5}/53.8 & \cellcolor{babyblue!15}\textcolor{tabblue}{58.7}/\textcolor{tabblue}{69.5} & \textcolor{tabblue}{65.5}/\textcolor{tabblue}{77.5} & \cellcolor{babyblue!15}\textcolor{tabblue}{56.6}/\textcolor{tabblue}{66.9} \\
PC2WF  & 11.9/26.8 & \cellcolor{babyblue!15}43.2/54.3 & 58.5/65.2 & \cellcolor{babyblue!15}37.9/48.8 & 0.84/7.61 & \cellcolor{babyblue!15}7.68/23.3 & 23.0/40.4 & \cellcolor{babyblue!15}10.5/23.8 \\ \midrule[0.5pt]
\textbf{Ours} & \textcolor{tabred}{\bfseries 91.3}/\textcolor{tabred}{\bfseries 92.2} & \cellcolor{babyblue!15}\textcolor{tabred}{\bfseries 93.4}/\textcolor{tabred}{\bfseries 93.9} & \textcolor{tabred}{\bfseries 94.4}/\textcolor{tabred}{\bfseries 94.8} & \cellcolor{babyblue!15}\textcolor{tabred}{\bfseries 93.0}/\textcolor{tabred}{\bfseries 93.6} & \textcolor{tabred}{\bfseries 76.8}/\textcolor{tabred}{\bfseries 84.7} & \cellcolor{babyblue!15}\textcolor{tabred}{\bfseries 80.6}/\textcolor{tabred}{\bfseries 87.1} & \textcolor{tabred}{\bfseries 83.9}/\textcolor{tabred}{\bfseries 89.5} & \cellcolor{babyblue!15}\textcolor{tabred}{\bfseries 80.4}/\textcolor{tabred}{\bfseries 87.1} \\ \bottomrule[0.8pt]
\end{tabular}
}
    \end{subtable}\hfill
    \begin{subtable}[t]{0.45\textwidth}
    \caption{\textbf{Wireframe Edit Distance} (WED) of the reconstructed \textbf{wireframes}. We report the number of operations (Num) and the editing distances in meters (Dist).}\label{tab:res:wed}
    \centering
\tiny
{\def\arraystretch{1.05}\tabcolsep=0.52em 
\begin{tabular}{l|cccccccc}
\toprule[0.8pt]
\multicolumn{1}{c}{} &  \multicolumn{2}{c}{(WED) \textbf{+vertex}} & \multicolumn{2}{c}{(WED) \textbf{+edge}} & \multicolumn{2}{c}{(WED) \textbf{-edge}} & \multicolumn{2}{c}{(WED) \textbf{Total}} \\ \cmidrule(lr){2-9} 
\multicolumn{1}{c}{\multirow{-3}{*}{Method}}  & Num. & \cellcolor{babyblue!15}Dist & Num & \cellcolor{babyblue!15}Dist & Num & \cellcolor{babyblue!15}Dist & Num & \cellcolor{babyblue!15}Dist \\ \midrule[0.8pt]

line2Surf.  & 1.012 & \cellcolor{babyblue!15}13.78 & 6.223 & \cellcolor{babyblue!15}35.76 & 9.427 & \cellcolor{babyblue!15}48.77 & 16.66 & \cellcolor{babyblue!15}98.31 \\
PolyFit & 1.681 & \cellcolor{babyblue!15}3.170 & 4.811 & \cellcolor{babyblue!15}21.41 & {0.969} & \cellcolor{babyblue!15}{5.285} & \textcolor{tabblue}{7.463} & \cellcolor{babyblue!15}\textcolor{tabblue}{29.86} \\
PC2WF &   5.216 & \cellcolor{babyblue!15}3.445 & 17.01  & \cellcolor{babyblue!15}87.94 & 4.622 & \cellcolor{babyblue!15}33.38 & 26.84  & \cellcolor{babyblue!15}124.8 \\ \midrule[0.5pt]
\textbf{Ours}  & {0.766} & \cellcolor{babyblue!15}{1.810} & {2.880} & \cellcolor{babyblue!15}{11.49} & 1.655 & \cellcolor{babyblue!15}14.03 & \textcolor{tabred}{\bfseries 5.301} & \cellcolor{babyblue!15}\textcolor{tabred}{\bfseries 27.33} \\ \bottomrule[0.8pt]
\end{tabular}
}

    \end{subtable}
    \vspace{10pt}
    \caption{Precision/Recall and Wireframe Edit Distance results after post-processing}
    \label{tab:res:PRandWED}
    \vspace{-10pt}
\end{table}

Tab.~\ref{tab:res:postprocess} shows a  fair comparison to line2Surf and PolyFit, where all the output meshes are post-processed into cleaner wireframes (see Fig.~\ref{fig:res:baselines} for some examples of the post-processed results; see Fig.~\ref{fig:mtd_overview} (B2,B3) and supplementary materials for examples of direct outputs from different methods). 
we report the number of vertex/edges, and the precision/recall of the predicted junctions/wireframes on the results after post-processing. 
We do not compare to Line3Dpp in this case since it outputs a nonstructural line cloud. The results show that our method outperforms all the baselines on building wireframe reconstruction. 

Note that, PolyFit shows visually comparable results to ours in Fig.~\ref{fig:res:baselines}, but its accuracy is much lower as shown in Tab.~\ref{tab:res:postprocess}. 
The reason is that in PolyFit, the junction positions are determined by the intersections among the \emph{estimated} planes, and their $L_2$ distance to the ground-truth junctions can be large even though the overall shape looks alike. 
At the same time, redundant faces will be generated to satisfy the watertight hard constraint in PolyFit.

%
Tab.~\ref{tab:res:wed} shows the wireframe edit distance for different methods, where our method achieves the least number of editing operations and the smallest editing distances.
Fig.~\ref{fig:res:baselines} shows a qualitative comparison of different baselines. 
We can see that line3Dpp can indeed provide more abstracted line clouds, but they are still far from clean wireframe models. 
Line2Surf can robustly recover planes from the line cloud (from Line3Dpp), but it is not robust to the noise. 
PC2WF is trained on point clouds from mechanical objects and furniture, which are likely to have a large domain gap to the rooftop structures. Therefore, the wireframes constructed by PC2WF can only recover the walls in building point clouds. Moreover, we also observe that the point clouds stemming from our scenes are more noisy (though they are accurate enough) than the point clouds that PC2WF is trained on. This can also lead to the less satisfactory results that PC2WF obtains.
PolyFit is a powerful method for building reconstruction that is robust to noisy point clouds. However, PolyFit can be computationally costly when the input point cloud is too dense since the algorithm involves integer linear programming. As a comparison, our method can achieve visually comparable and quantitatively better results in a much more efficient way. For example, on average it takes our method 0.9s to infer a building wireframe while it takes PolyFit 45.6s to optimize a building mesh (see Tab.~\ref{tb:res:baseline}).
We show more results of our reconstructed wireframes in the supplementary materials.

\setlength{\columnsep}{2pt}%
\setlength{\intextsep}{0pt}%
\begin{wrapfigure}{r}{0.6\linewidth}
\centering
    \begin{overpic}[trim=2cm 1.5cm 1cm 1.2cm,clip,width=1\linewidth,grid=false]{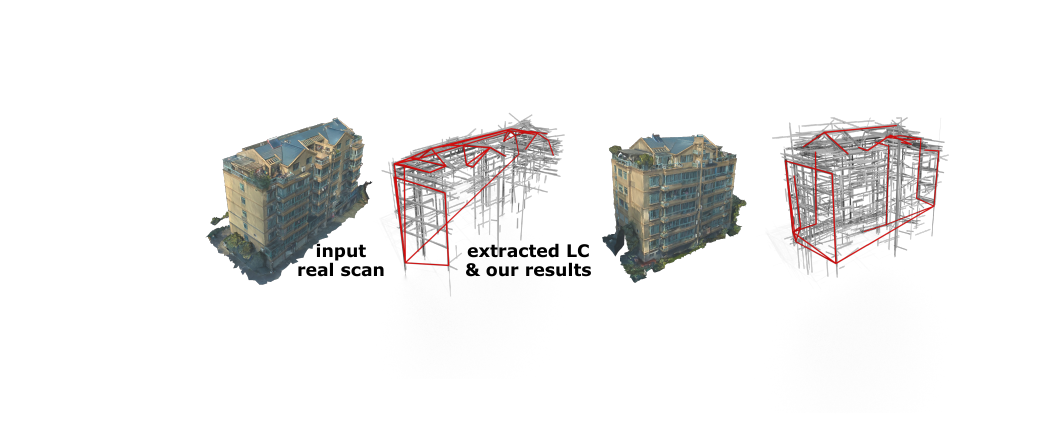}
    \end{overpic}
    \vspace{-3pt}
    \caption{\textbf{Two real-data examples (without finetuning)}: we overlay our reconstructed wireframe (\emph{red}) on top of the extracted line cloud (\emph{gray}).}\label{fig:rebuttal:eg_real}
    \vspace{-5pt}
\end{wrapfigure}

Fig.~\ref{fig:rebuttal:eg_real} shows some preliminary but reasonable results on two real-world noisy scans without finetuning.
One of the main challenges of our task is the lack of large-scale real-world buildings paired with clean and complete wireframes (e.g., manually created by artists). Adapting existing datasets is almost as hard as designing a new one as discussed in Sec.3.2. 
Inspired by PC2WF where synthetic point clouds are generated for training and testing, we therefore justify our LC2WF on synthetic dataset.
We believe our LC2WF can be fine-tuned on future real-world datasets to get better performance.


\vspace{-15pt}
\section{Conclusion, Limitation \& Future Work}
In this work, we present the first learning-based solution for building wireframe reconstruction from line clouds, which can be efficiently extracted from multi-view images. 
We construct a synthetic dataset, BuildingWF, containing multi-view images of 3.6K buildings and the corresponding ground-truth wireframe models.
The key component of our method is a Line-Patch Transformer which can be used for junction and connectivity prediction from line patches, a group of neighboring line segments that potentially encode the contour information of the underlying building.
Our method outperforms multiple state-of-the-art building reconstruction methods on both accuracy and efficiency.

Our method still has some limitations. 
For example, we assume the input multi-view images cover the overall region of the underlying buildings, and we expect to extract a building wireframe from a noisy but relatively complete line clouds. Therefore, no prior knowledge or extra regularizers are investigated to complete a wireframe from a partial line cloud with large missing regions. We would like to leave it as future work to investigate wireframe reconstruction from partial line clouds.
Moreover, in this work we do not investigate how to convert a wireframe into a watertight mesh. We believe it would be interesting to try to learn face information from line patches as well, which we leave as future work.
\vspace{-10pt}
\paragraph{Acknowledgments} The authors thank the anonymous reviewers for their valuable comments and suggestions. 
We would like to acknowledge support from the SDAIA-KAUST Center of Excellence in Data Science and Artificial Intelligence.
We thank \emph{Zhenbang He} and \emph{Yue Qian} for their helpful suggestions.

\clearpage
\citestyle{bmvc2k}
\bibliography{egbib}
\end{document}